\documentclass[letterpaper, 10 pt, conference]{ieeeconf}  % Comment this line out if you need a4paper

\IEEEoverridecommandlockouts                              % This command is only needed if 
\title{\LARGE \bf
\ourmethod{}: Fast Motion Planning Using Diffusion Models Based on \\Key-Configuration Environment Representation}

\author{Mingyo Seo$^{1*}$, Yoonyoung Cho$^{2*}$, Yoonchang Sung$^{1}$, Peter Stone$^{1,3}$, Yuke Zhu$^{1}$$^{\dagger}$, and Beomjoon Kim$^{2}$$^{\dagger}$
\thanks{$^{1}$The University of Texas at Austin,}
\thanks{$^{2}$Korea Advanced Institute of Science and Technology, $^{3}$Sony AI,}
\thanks{$^{*}$Equal contribution, $^{\dagger}$Equal advising.} 
\thanks{Correspondance: {\tt\small beomjoon.kim@kaist.ac.kr}}
}

\let\labelindent\relax
\DeclareUnicodeCharacter{0301}{\'{e}}
\usepackage{times}
\usepackage{xcolor}

\usepackage{amsfonts}
\usepackage{amsmath}
\usepackage{amssymb}

\usepackage{cuted}
\usepackage{capt-of}

\usepackage{graphicx}
\usepackage{grffile}
\usepackage{booktabs}
\usepackage{hyperref}
\usepackage{dsfont}
\usepackage{svg}

\usepackage{setspace}

\usepackage[font=footnotesize, labelfont=bf]{caption}

\usepackage{enumitem}
\usepackage[backend=biber,
            hyperref=true,
            url=false,
            isbn=false,
            doi=false,
            backref=false,
            style=ieee,
            citestyle=numeric-comp,
            sorting=none,
            block=none]{biblatex}
\usepackage[font=footnotesize]{caption}

\usepackage{algpseudocode}
\usepackage{algorithmicx}
\usepackage{algorithm}
\usepackage{dblfloatfix}

\setlist[itemize]{leftmargin=*}

\renewcommand{\bibfont}{\small}
\newcommand{\etal}{\textit{et al}.}

\newcommand{\ourmethod}{PRESTO}
\newcommand{\diffuser}{SceneDiffuser}
\newcommand{\mpd}{MPD}
\newcommand{\trajopt}{TrajOpt}
\newcommand{\birrt}{Bi-RRT}

\newcommand{\task}[1]{\textbf{\textit{#1}}}
\newcommand{\level}[1]{{\textit{Level #1}}}
\newcommand{\claim}[1]{{\textit{Claim #1}}}

\algblock{ParFor}{EndParFor}
\algnewcommand\algorithmicparfor{\textbf{parallel for}}
\algnewcommand\algorithmicpardo{\textbf{do}}
\algnewcommand\algorithmicendparfor{\textbf{end\ parallel}}
\algrenewtext{ParFor}[1]{\algorithmicparfor\ #1\ \algorithmicpardo}
\algrenewtext{EndParFor}{\algorithmicendparfor}
\begin{document}
\maketitle
\thispagestyle{empty}
\pagestyle{empty}

%===============================================================================

\begin{abstract}
We introduce a learning-guided motion planning framework that generates seed trajectories using a diffusion model for trajectory optimization. 
Given a workspace, our method approximates the configuration space (C-space) obstacles through an environment representation consisting of a sparse set of task-related key configurations, which is then used as a conditioning input to the diffusion model.
The diffusion model integrates regularization terms that encourage smooth, collision-free trajectories during training, and trajectory optimization refines the generated seed trajectories to correct any colliding segments.
Our experimental results demonstrate that high-quality trajectory priors, learned through our C-space-grounded diffusion model, enable the efficient generation of collision-free trajectories in narrow-passage environments, outperforming previous learning- and planning-based baselines. 
Videos and additional materials can be found on the project page: \url{https://kiwi-sherbet.github.io/PRESTO}.
\end{abstract}

%===============================================================================

\section{Introduction}
Motion planning involves finding a smooth and collision-free path in a high-dimensional configuration space (C-space). 
Classical motion planning algorithms typically use either sampling-based methods~\cite{lavalle1998rapidly, lavalle2001rapidly, kavraki1996probabilistic} or optimization-based methods~\cite{ratliff2009chomp, schulman2014motion} to address motion planning across various domains. 
However, in high-dimensional C-spaces with narrow passages, sampling-based methods incur high computational costs due to large search spaces and small volume of solutions.
While optimization-based methods can serve as an alternative, such methods are sensitive to initialization and may become stuck in local optima, often failing to find a feasible path. 
Consequently, both approaches have limitations
when dealing with complex motion planning problems under restricted computational resources.

Recent works leverage generative models to directly learn trajectory distributions instead~\cite{janner2022planning, huang2023diffusion, carvalho2023motion}.
By casting motion planning as sampling from a learned distribution, these models can efficiently generate trajectories within a consistent computational budget.
However, they often struggle to generalize to new, complex C-spaces, resulting in high collision rates in the generated trajectories, because most of these approaches use the workspace as input to neural networks instead of the C-space.
Instead, we propose representing the environment in terms of key configurations~\cite{kim2019adversarial}, a sparse set of task-related configurations from prior motion planning data. 
The resulting model no longer needs to learn a generalizable mapping between workspace and C-space obstacle representations, improving generalization and reducing training complexity.

\begin{figure}
	\centering
	\includegraphics[width=0.98\linewidth]{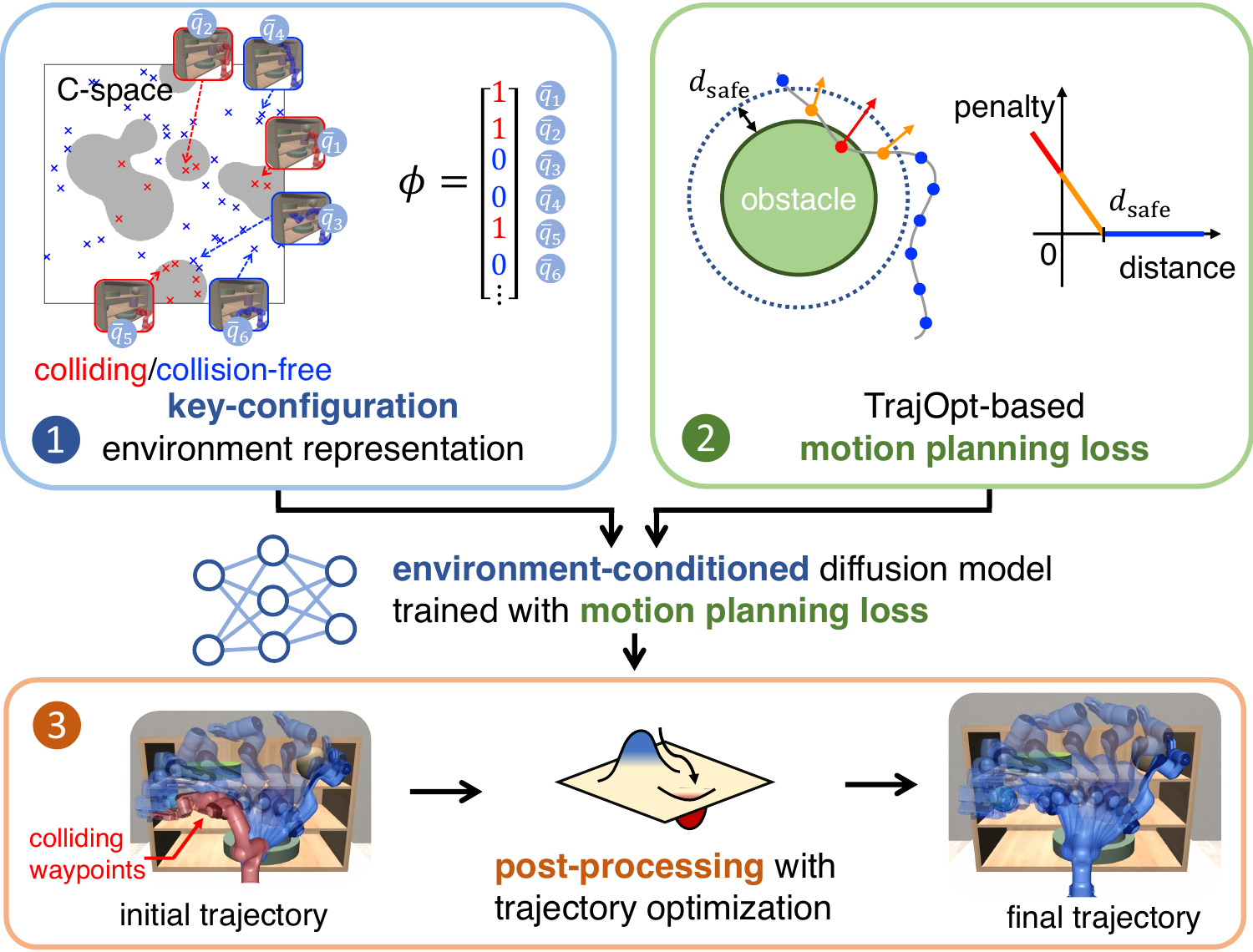}
	\caption
	{
        \textbf{Overview of \ourmethod{}.}
        \ourmethod{} aims to generate collision-free trajectories in complex, unseen C-spaces. 
        First, we approximate these C-spaces using key configurations from prior data and generate trajectories based on this representation. 
        A conditional diffusion model, trained with a motion planning loss, provides initial solutions that are subsequently refined through trajectory optimization.
    	}
     \vspace{10pt}
	\label{fig:header}
\end{figure}

Another challenge is designing a training objective for generative models tailored to motion planning.
% \TODO{BK's comment: the use of diffusion model is introduced rather }
Existing diffusion models for motion planning use DDPM-based losses~\cite{ho2020denoising,janner2022planning,carvalho2023motion} that focus on reconstruction quality~\cite{saharia2022imagen}.
However, this reconstruction objective does not account for underlying task constraints, resulting in degraded performance on tasks that require precise outputs and complex constraints~\cite{giannone2023aligning}.
To overcome this, we incorporate TrajOpt-inspired motion-planning costs~\cite{schulman2014motion} directly into the training pipeline of a diffusion model, minimizing trajectory-optimization costs associated with collision avoidance and trajectory smoothness.
As a result, the model learns to generate smooth and collision-free trajectories.
Further, to ensure that generated trajectories satisfy hard constraints such as collision avoidance, we feed the diffusion model's outputs as initial solutions for subsequent trajectory optimization.

\begin{figure*}
    \vspace{4pt}
	\centering
	\includegraphics[width=0.86\linewidth]{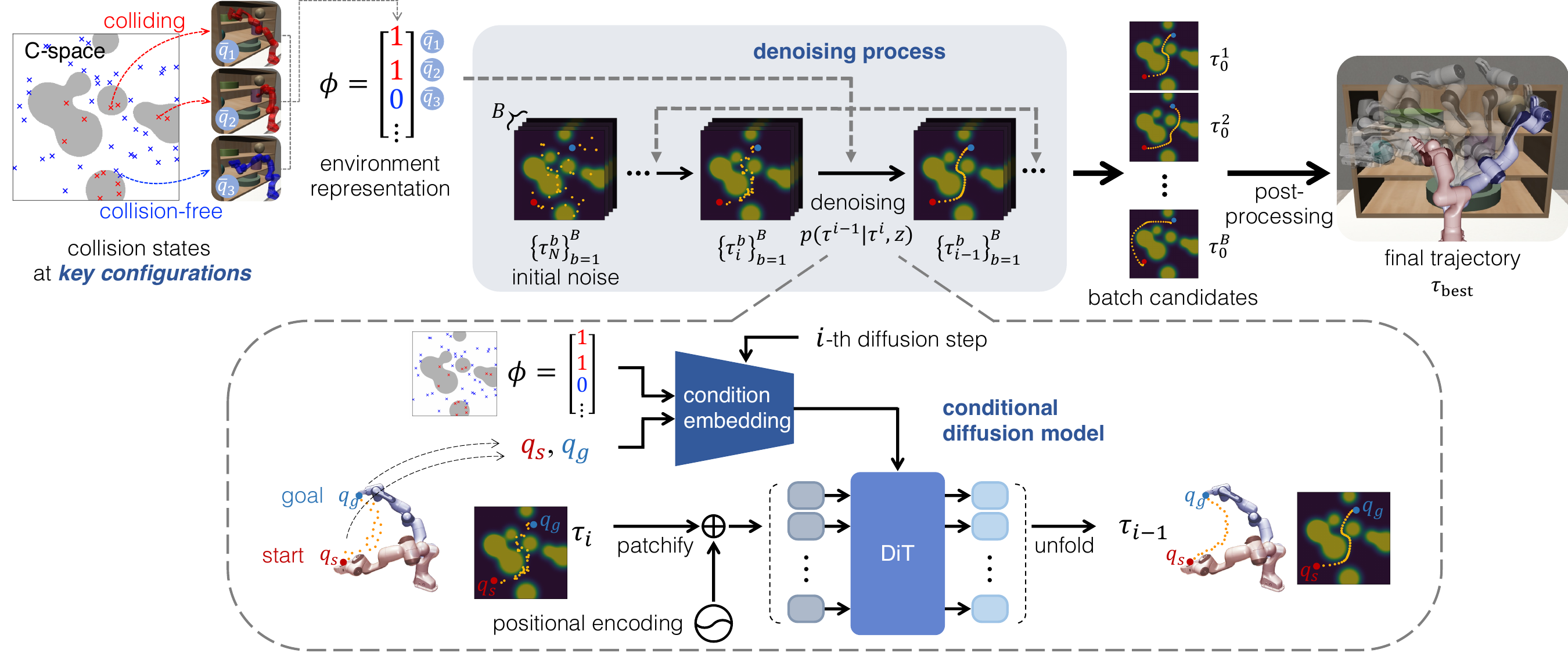}
	\caption
	{
        \textbf{Trajectory generation pipeline of \ourmethod{}.}
        We obtain the environment representation for an unseen problem by checking the collision states at the key configurations used during training.
        Using the trained conditional diffusion model, we generate multiple trajectories conditioned on this representation and then select the least-colliding trajectory after post-processing.
        }
    \label{fig:pipeline}
\end{figure*}

We name our unified framework \ourmethod{} (\underline{P}lanning with Environment \underline{Re}presentation, \underline{S}ampling, and \underline{T}rajectory \underline{O}ptimization). 
Figure~\ref{fig:header} summarizes our framework:
1) an environment representation based on key configurations (blue block); 
2) a training pipeline for a diffusion model that directly integrates motion-planning costs (green block); and 
3) a diffusion-based sampling-and-optimization framework for motion planning, where the diffusion model provides initial trajectories for trajectory optimization (orange block).
We evaluate \ourmethod{} in simulated environments where a robot operates in a fixed scene populated with randomly shaped and arranged objects.
The results show that the synergy between the high-quality trajectory priors generated by our diffusion model and the trajectory optimization post-processing efficiently generates collision-free trajectories in narrow passages within a limited computational budget.
\section{Related Work}

Several approaches learn to guide motion planning by complementing motion planners with learned models for collision checking~\cite{huh2016learning, das2020learning, danielczuk2021object, kew2020neural, zhi2022diffco, murali2023cabinet} or for sampling promising configurations~\cite{arslan2015machine, iversen2016kernel, zhang2018learning, kuo2018deep, ichter2018learning, kumar2019lego, yu2021reducing, kim2018aaai}.

A series of work has focused on learning collision checkers to expedite motion planning.
One body of work progressively learns collision models in a given environment as Gaussian mixtures~\cite{huh2016learning} or kernel perceptrons~\cite{das2020learning,zhi2022diffco}, while others use deep neural network models trained on large offline datasets.
Kew et al.~\cite{kew2020neural} train a neural network to estimate the clearance of robot configurations in a fixed environment. Danielczuk et al.~\cite{danielczuk2021object} generalize collision predictions to diverse environments by training a neural network to estimate collision states from object and scene point clouds.
Recently, Murali et al.~\cite{murali2023cabinet} extend neural collision detectors to a partially observable setting. 
Our approach can integrate with these collision-checking methods to further enhance planning speed.

Another line of work focuses on sampling promising configurations. 
Some methods derive explicit distributions via kernel density estimation~\cite{arslan2015machine, iversen2016kernel}, while others use neural networks, in which sequence-based models such as LSTMs~\cite{kuo2018deep} or a rejection-sampling policy~\cite{zhang2018learning} learn the distribution of collision-free configurations based on the history of collision states at previously sampled configurations.
Alternatively, recent works propose using generative models to govern the sampling process.
Qureshi et al.~\cite{qureshi2020motion} map point-cloud inputs to the next sampling configuration, stochastically applying dropout in the interim layers to generate diverse samples; Yu et al.~\cite{yu2021reducing} use Graph Neural Networks to identify promising regions of a roadmap; and Ichter et al.~\cite{ichter2018learning} along with Kumar et al.~\cite{kumar2019lego} employ Conditional Variational Autoencoders (CVAEs) to sample task-related configurations from latent spaces. 
While these methods generate individual or sets of joint configurations, recent works propose using diffusion models to learn trajectory sampling distributions to accelerate motion planning.

Diffuser~\cite{janner2022planning} first explored diffusion models for generating trajectories across various start and goal configurations in a fixed environment. 
Subsequent work aims to generalize to unseen environments using test-time guidance~\cite{saha2024edmp, carvalho2023motion}, but struggles with large environmental variations due to significant mismatches in learned trajectory distributions. 
Recent studies~\cite{ajay2022conditional} address this issue by conditioning diffusion models on environmental conditions. 
Notably, Huang et al.~\cite{huang2023diffusion} and Xian et al.~\cite{xian2023chaineddiffuser} condition on point-cloud representations to sample motion plans across diverse environments. 
However, these models still face challenges in generalizing to a new C-space because they rely on workspace inputs, even though motion planning occurs in the C-space.

Prior work explores alternative representations for motion planning, focusing on C-space grounded approaches~\cite{jetchev2013fast, kim2019adversarial}.
These methods approximate complex C-spaces from past problems to find collision-free trajectories in unseen environments. 
In particular, our work incorporates a key-configuration representation inspired by Kim et al.~\cite{kim2019adversarial}, which utilizes collision states at task-related key configurations to enhance model generalization across varying C-spaces.
\vspace{-2pt}
\section{Problem Description}
\label{sec:prob}

Let \(\mathcal{C}\) be a \(d\)-dimensional C-space, which is divided into two subspaces: \(\mathcal{C}_o\) representing C-space obstacles, and \(\mathcal{C}_f = \mathcal{C} \setminus \mathcal{C}_o\) representing the collision-free C-space. We denote the robot's configuration as a \(d\)-dimensional vector \(q \in \mathcal{C}\). A trajectory is represented as a sequence of waypoint configurations \(\tau = (q_0, q_1, \ldots, q_T)\). 
Given the start configuration $q_s$ and goal configuration $q_g$, the objective of motion planning is to find a collision-free path \(\tau \in \mathcal{C}_f\) from $q_s$ to $q_g$.

In this work, we assume that we are provided with a dataset $\mathcal{D}=\{\mathcal{D}_m\}_{m=1}^{M}$ obtained from solving $M$ past planning problems, where each data point $\mathcal{D}_m=\{q_s, q_g, \tau, \mathcal{G}\}$ consists of a start configuration $q_s$, a goal configuration $q_g$, a trajectory $\tau$, and an environment geometry $\mathcal{G}$.
% .s with varying $\mathcal{C}_o$ and $\mathcal{C}_f$.
% where background fixtures remain consistent across problems and objects vary randomly in size, shape, and location.
We assume consistent environment fixtures but varying object shapes and locations across problems.
We use an optimization-based planner to compute ground-truth trajectories for training data.
Our goal is to develop a generative model that provides a good initial solution for trajectory optimization, even in environments with unseen obstacles and their arrangements.
\section{Method}

\ourmethod{} comprises of three key components, illustrated in Figure~\ref{fig:header}.
First, we generate a set of key configurations and their collision states from the motion planning dataset.
Based on the resulting representation, we train a conditional diffusion model that incorporates motion-planning costs to guide the model toward smooth and collision-free trajectories.
Finally, we feed the trajectories generated by the diffusion model to trajectory optimization.
Each component of our framework is detailed in the following sections.

\begin{algorithm}[b]
\caption{Key-Configuration Selection.}
\label{algo:key-config}
\begin{algorithmic}[1]
\scriptsize

\Require C-space/Workspace separation distance $d_q^{\text{min}},d_x^{\text{min}}$,
\Statex Collision proportion bound $c$, Motion plan dataset $\mathcal{D}$,
\Statex Number of key configurations $K$
\Ensure Key configurations $\{ \bar{q}^{k} \}_{k=1}^{K}$

\Statex // Initialization
\State $\{\overline{q}\} \gets \emptyset$ \Comment{Initialize the key configuration buffer}

\Statex // Sampling and selecting key configurations
\While{$|\{\overline{q}\}| < K$}
    \State $\{\tau, q_s, q_g, \mathcal{G}\} \sim \mathcal{D}$
    \Comment{Sample a motion plan instance}
    \State $q \sim \tau$
    \Comment{Sample a configuration}
    \State $d_q = \texttt{MinCSpaceDistance}(\{\overline{q}\} \cup \{q\})$
    \State $d_x = \texttt{MinWorkspaceDistance}(\{\overline{q}\} \cup \{q\})$
    \State $p_c = \frac{1}{M} \sum_{m=1}^M \texttt{EnvCollision}(q, n) $
    \If{$d_q > d_q^{\text{min}}$ \textbf{and}
    $d_x > d_x^{\text{min}}$ \textbf{and}
    $p_c \in (c, 1-c)$
    }
    \State $\{\overline{q}\} \leftarrow \{\overline{q}\} \cup \{q\}$
    \EndIf
\EndWhile

\State \Return $\{\overline{q}\}$
\end{algorithmic}
\end{algorithm}

\subsection{Environment Representation}
\label{subsec:env-representation}
We represent the environment as an approximation of its C-space using a collection of key configurations selected from the dataset. 
We denote the set of key configurations as \(\{\overline{q}^k\}_{k=1}^K\), where the number of key configurations \(K\) determines the resolution of the environment's C-space approximation\footnote{A larger $K$ increases the resolution of the C-space approximation, but it also raises the computational overhead at query time.}.
For each motion planning problem, we compute the environment representation $\phi \in \{0, 1\}^{K}$ as a binary vector that specifies the collision states of each key configuration. In our setups, we use $K=1025$.

Algorithm~\ref{algo:key-config} describes our procedure for generating key configurations, which is a modified version of the algorithm originally proposed by Kim et al.~\cite{kim2018aaai}.
It takes the dataset $\mathcal{D}$ and the hyperparameters ${d_q^{\text{min}}, d_x^{\text{min}}, c, K}$. 
Here, $d_q^{\text{min}}$ denotes the minimum C-space distance between key configurations, 
$d_x^{\text{min}}$ represents the minimum workspace distance for end-effector tips, 
and $c$ is the bound on the proportion of environments where a key configuration is occupied.

We initialize the key configuration set as an empty set (line 1) and sample new configurations until the target size is reached (lines 2-11).
At each step, a configuration is uniformly sampled from $\mathcal{D}$ (lines 3-4) and filtered based on three conditions (lines 5-10).
To avoid duplicates, we ensure the new configuration is sufficiently distant from existing ones in both C-space and workspace (lines 5-6), while also limiting the proportion of collision states across different environments to prioritize informative configurations\footnote{By limiting the proportion of collision states, we filter out configurations that never result in collisions, as they provide no meaningful information about the environment
.} (line 7). 
If all criteria are met, the configuration is added to the buffer (lines 8-10).
This process generates key configurations that effectively capture task-relevant C-space regions for motion planning.

\subsection{Training Conditional Diffusion Models\label{sec:train}}
\label{subsec:trajectory}

\paragraph{Model Architecture}
Figure~\ref{fig:pipeline} (bottom) illustrates our model architecture, which is based on the Diffusion Transformer (DiT)~\cite{Peebles2022DiT}. 
Our model takes as inputs the current diffusion step $i$, the noisy trajectory at the $i$-th step $\tau_{i}$, the environment representation $\phi$, and the start and goal joint configurations $q_{s}$ and $q_{g}$.
We use v-prediction~\cite{salimans2022vpred} during inference to enhance sample quality~\cite{saharia2022imagen}.

To process the inputs, the trajectory $\tau_{i}$ is first patchified and tokenized by an MLP, as in DiT~\cite{Peebles2022DiT}.
The diffusion step $i$ and the start and goal configurations $q_s$ and $q_g$ are mapped to high-dimensional frequency embeddings~\cite{Peebles2022DiT} to capture small changes.
The embedded trajectory patches are given as input tokens to the transformer,
while $i$, $\phi$, $q_s$, and $q_g$ are incorporated as conditioning inputs to align sampled trajectories with the current scene and endpoint constraints.

The DiT comprises six transformer blocks to process the trajectories, where each block incorporates Adaptive Layer Normalization (AdaLN)~\cite{Peebles2022DiT} that transforms the output features based on the conditioning inputs. In AdaLN, a separate MLP maps conditioning inputs to the transform parameters as ${\gamma, g, b} = \text{MLP}({i, \phi, q_s, q_g})$, applied to the output $x$ as $\hat{x} = x + \gamma \odot ((1 + g) \odot \text{LN}(x) + b)$, where $\text{LN}(x)$ denotes layer normalization and $\odot$ denotes element-wise multiplication. 
This allows the model to adjust its output based on the current scene and the diffusion iteration.

\paragraph{Training with Motion-Planning Costs}
Our training objective includes three terms: \textit{Diffusion Loss}, \textit{Collision Loss}, and \textit{Smoothing Loss}.  
While \textit{Diffusion Loss} implements the standard reconstruction objective used in diffusion models, we add \textit{Collision Loss} and \textit{Smoothing Loss} to encourage the model to learn the motion planning constraints.
\begin{itemize}
\item \textbf{\textit{Diffusion Loss}:} This term $\mathcal{L}_{\text{diffusion}}$ represents the standard loss function used for training conditional diffusion models based on the DDPM framework.
\item \textbf{\textit{Collision Loss}:} Inspired by \trajopt{}~\cite{schulman2014motion}, this term encourages the model to generate trajectories that maintain a safe distance from objects and other links. It is defined as\\
\vspace{-8pt}
\begin{equation*}
    \mathcal{L}_{\text{coll}}=
\sum_{i,j} |d_\text{safe} - \text{sd}(\mathcal{A}_i, \mathcal{O}_j)|^+
+ \sum_{i \neq j} |d_\text{safe} - \text{sd}(\mathcal{A}_i, \mathcal{A}_j)|^+,
\end{equation*}
\vspace{-10pt}
\\where $|\cdot|^+ = \text{max}(\cdot, 0)$ and $\text{sd}(\cdot) = \text{dist}(\cdot) - \text{penetration}(\cdot)$ with $d_\text{safe} = 0.01$ m as the safety margin of the hinge loss. This loss is summed over each robot link $\mathcal{A}_i$ and object $\mathcal{O}_j$ based on the swept volume of the trajectory. 
The first term accounts for the collision between the $i$-th link $\mathcal{A}_i$ and the $j$-th obstacle $\mathcal{O}_j$, while the second term denotes self-collision among different links $\mathcal{A}_i$ and $\mathcal{A}_j$, where $i \neq j$.
\item \textbf{\textit{Smoothing Loss}:} This term penalizes the L2-norm between adjacent configurations, defined as $ \mathcal{L}_{\text{smooth}} = \sum_{t} \left\| q_{t} - q_{t-1} \right\|^2$. It regulates the distances between consecutive configurations to encourage shorter and smoother trajectories for the robot.
\end{itemize}
We use a weighted sum of the loss terms to train our model: $\mathcal{L} = w_1 \mathcal{L}_{\text{diffusion}} + w_2 \mathcal{L}_{\text{coll}} + w_3 \mathcal{L}_{\text{smooth}}$, with $w_1=1.0$, $w_2=0.05$, and $w_3=0.005$. While the model is not highly sensitive to these values, $w_2$ is kept small for stable training.

\begin{algorithm}[b]
% \SetKwInOut{Parameter}{parameter}
\caption{\ourmethod{} Trajectory Generation.}
\label{algo:pipeline}
\begin{algorithmic}[1]
\scriptsize

\Require Start/Goal configurations $\{q_s,q_g\}$, Environment $\mathcal{G}$,
\Statex Key configurations $\{ \bar{q}^{k} \}_{k=1}^{K}$, Diffusion model $\mu_{\theta}(\cdot)$,
\Statex Noise schedule $\{ \sigma_i \}_{i=1}^{N}$

\Ensure Output trajectory $\tau_{\text{best}}$

\Statex // Computing environment representation
\State $\phi = \texttt{CheckCollision}(\mathcal{G}, \{ \bar{q}^{k} \}_{k=1}^{K})$

\Statex // Batched denoising with reverse process
\State $\{\tau^b_N \sim \mathcal{N}(\textbf{0}, \textbf{I})\}_{b=1}^B $ \Comment{Sample a batch of initial trajectories}
\ParFor{$b = 1, ..., B$}
\For{$i = N, ..., 1$}
\State $\tau_{i-1}^b\sim\mathcal{N}\big(\mu_\theta(\tau_i^b, \phi, i), \sigma_i\big)$
\State $q_0 \gets q_s$, $q_T \gets {q_g}$ \Comment{Apply endpoint constraints $q_s, q_g$}
\EndFor
\EndParFor
\State $\tau_{\text{seed}}$ = $\{\tau_0^b\}_{b=1}^{B}$ \Comment{Batch-sized sampled trajectories}

\Statex // Post-processing
\State $\tau_{\text{opt}} = \texttt{TrajectoryOptimization} (\tau_{\text{seed}})$
\State $\tau_{\text{best}} = \texttt{BestTrajectory} (\tau_{\text{opt}})$

\State \Return $\tau_{\text{best}}$
\end{algorithmic}
\end{algorithm}

\subsection{Trajectory Generation}
Figure~\ref{fig:pipeline} (top) provides an overview of our trajectory generation process. 
The inputs ${q_s, q_g, \phi}$ specify the motion planning problem, where $q_s$ and $q_g$ are the start and goal configurations, and $\phi$ is the environment representation from the key configurations' collision states. 
During denoising, we use batch sampling to leverage GPU parallelization, enhancing the likelihood of finding collision-free trajectories among the diffusion model’s stochastic outputs.
Our sampling follows the Denoising Diffusion Implicit Model~\cite{song2020denoising}, which accelerates the process by using fewer denoising iterations during inference than during training.
We then apply trajectory optimization to post-process the sampled trajectories and select the one with the lowest collision cost. 

We detail our trajectory generation in Algorithm~\ref{algo:pipeline}. 
First, we compute the environment representation $\phi$ by checking the collision states of the key configurations ${\overline{q}}$ (line 1). 
Next, we initialize the denoising process with a batch of random noise $\tau_N$ sampled from an isotropic Gaussian distribution (line 2).
At each of the $N$ iterations, we predict a denoised trajectory $\tau_{i-1}$ from $\tau_i$, conditioned on $\phi$ and the current step $i$ (line~5), and then apply endpoint constraints to ensure connectivity between the start $q_s$ and the goal $q_g$, as in Diffuser~\cite{janner2022planning} (line~6).
The resulting trajectories $\tau_{\text{seed}}$ provide initialization for post-processing (line 9).
In our experiments, we denoise $B=4$ trajectories over $N=64$ iterations.

In the post-processing phase, we refine the sampled trajectories using a fixed number of trajectory optimization iterations~\cite{schulman2014motion} to address potential collisions (line 10). 
To accelerate this process, we employ cuRobo~\cite{sundaralingam2023curobo}, which can batch-process the trajectories while eliminating data exchange across devices.
Afterward, we select the trajectory with the fewest collisions (line 11).
\section{Experiments}
\label{sec:exp-setup}

\subsection{Experimental Setup}
\begin{figure*}
    \vspace{4pt}
    \centering
	\includegraphics[width=\linewidth]{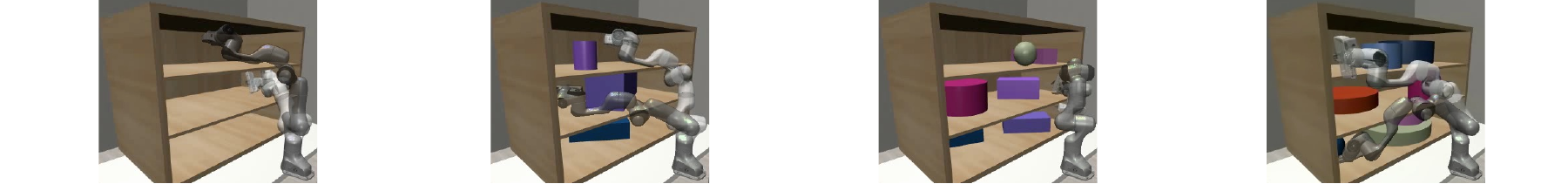}
	\includegraphics[width=\linewidth]{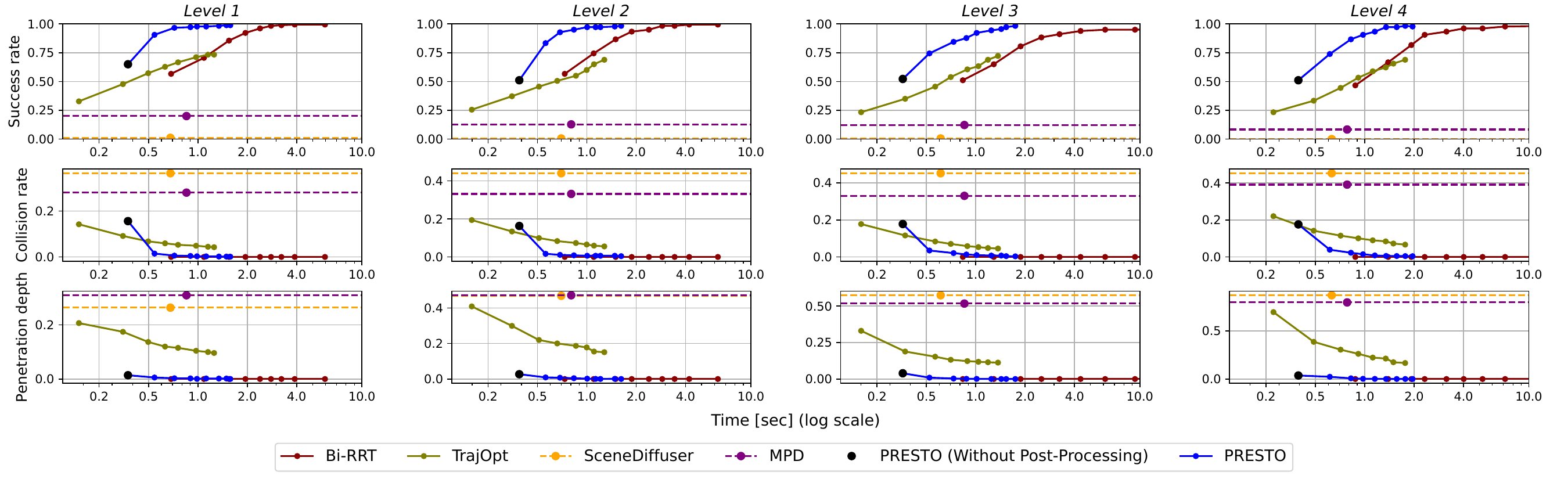}
	\caption
	{
        \textbf{Main results.}
        We report the success rate (\%), the collision rate (\%), and the penetration depth (m) across 180 problems. 
        (\textbf{Top}) The evaluation environments feature consistent 3-tier shelf fixtures with randomized object positions that vary across levels. 
        (\textbf{Bottom}) We show \ourmethod{}'s performance changes across domains and computational budgets compared to the baselines.
        }
    \label{fig:benchmark}
\end{figure*}

We evaluate our method on a motion planning task in which the Franka Emika \textit{Panda} robot arm~\cite{franka-panda} traverses a 3-tier shelf with various objects in simulation (Figure~\ref{fig:benchmark}, top).

\paragraph{Training Setup}
Our training domain consists of 5,000 environments, where 1-6 objects (cuboid, cylinder, sphere) are placed in random poses within each shelf slot. For each scene, we first sample collision-free initial and target joint positions within the workspace and then generate motion plans via cuRobo~\cite{sundaralingam2023curobo}. We collect a total of 50,000 environment-trajectory pairs (trajectory length $T = 1000$) annotated with key-configuration labels as in Algorithm~\ref{algo:key-config}.

\paragraph{Evaluation Setup}
While the evaluation domain is generated using a similar procedure, we partition the dataset into four difficulty levels, each consisting of 180 problems. 
This helps assess the performance of \ourmethod{} and the baselines on unseen scenes of varying complexity.

\begin{itemize}
    \item \textbf{\level{1}:} 
    The shelf is empty, as shown in Figure~\ref{fig:benchmark} (top left). 
    Although the environment remains consistent, the task is challenging due to the shelf’s non-convex workspace and the need to connect random start and goal configurations.
    \item \textbf{\level{2-3}:} Each slot contains one object for \level{2} and two objects for \level{3}, adding complexity to the C-space and requiring environment-conditional collision-free trajectory generation, as shown in Figure~\ref{fig:benchmark} (top center).
    \item \textbf{\level{4}:} Each slot contains 3-4 objects, as shown in Figure~\ref{fig:benchmark} (top right). 
    These environments are the most challenging due to narrow passages between obstacles, increased C-space complexity, and slower collision-checking and distance calculations among many objects.
\end{itemize}
We use the following metrics to evaluate the trajectories generated by \ourmethod{} and other motion planning methods:
\begin{itemize}
    \item \textbf{Success rate:} the percentage of collision-free trajectories within the batch. Higher is better.
    \item \textbf{Collision rate:} the average fraction of colliding segments in each trajectory.
    This metric reflects the likelihood that each joint configuration is collision-free, even if collisions occur elsewhere in the trajectory.
    Lower is better.
    \item \textbf{Penetration depth:} the average maximum penetration depth in each trajectory.
    This metric quantifies the severity of collision, measuring worst-case deviation from a collision-free trajectory.
    Lower is better.% as it indicates easier trajectory optimization.
\end{itemize}
To systematically evaluate how performance changes across different computational budgets,
we vary the number of optimization iterations during post-processing.
The \birrt{} baseline was evaluated on an AMD Ryzen 9 5900X and all other baselines were evaluated on an NVIDIA A5000.

\begin{figure*}
    \vspace{4pt}
	\centering
	\includegraphics[width=\linewidth]{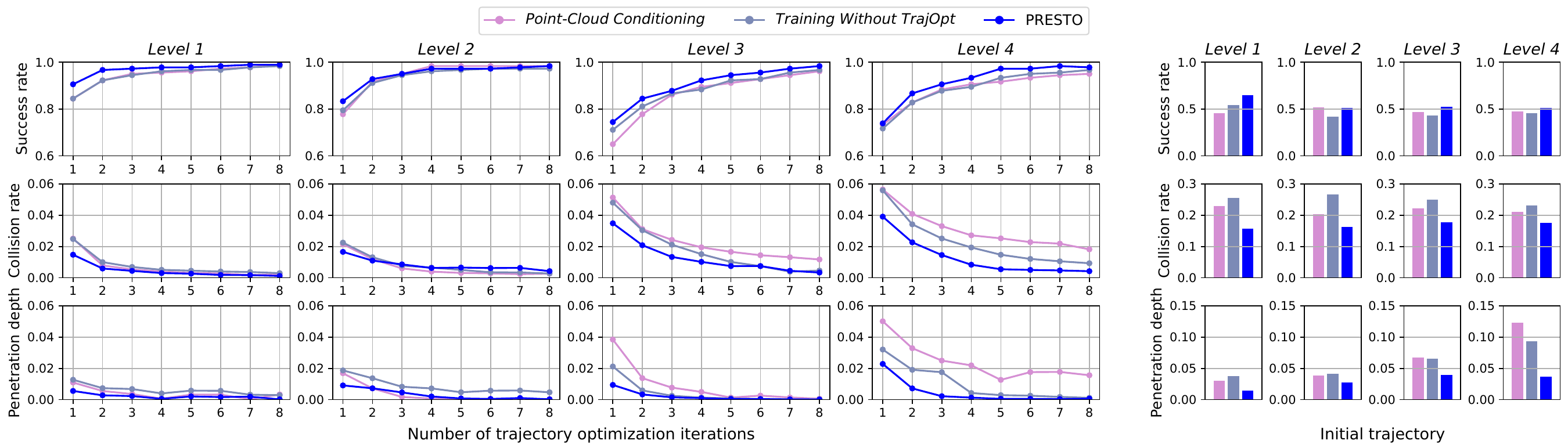}
	\caption
	{
    \textbf{Ablation study results.} We report the success rate (\%), the collision rate (\%), and the penetration depth (m) averaged across 180 problems for \ourmethod{} and the self-variant baselines. (\textbf{Left}) We show performance changes with varying post-processing iterations. (\textbf{Right}) We present the performance of trajectories directly generated by the diffusion models without post-processing.
}
    \label{fig:ablation}
\end{figure*}

\subsection{Quantitative Evaluation}
\label{sec:main_results}
In our experiments, we evaluate the following claims:
(\claim{1})
    Diffusion models using key-configuration representations generalize better to unseen environments than those using point-cloud representations.
(\claim{2})
    Incorporating motion-planning costs into diffusion-model training enables models to better learn task constraints like collision avoidance than when using only the reconstruction-based objective.
(\claim{3})
    By seeding trajectories, our method outperforms pure planners in computational efficiency and learning-based approaches in trajectory quality.

To validate our claims, we compare our model's performance against the following baselines, representative of each approach, as presented in Figure~\ref{fig:benchmark} (bottom).
\begin{itemize}
\item \textbf{\birrt{}:}
a pure planner based on LaValle et al.~\cite{lavalle2001rapidly}, specifically RRT-connect~\cite{kuffner2000rrt}.
\birrt{} searches bidirectionally by growing random trees from both start and goal configurations to find collision-free paths in an unknown C-space.
Although it is probabilistically complete, it empirically struggles with slow searches in narrow passages.
Since its success rate depends on the provided time, we evaluate its performance across different search timeouts.
\item \textbf{\trajopt{}:}
a pure planner from Schulman et al.~\cite{schulman2014motion} that optimizes trajectories using a hinge penalty for collisions and configuration distances.
The optimization scheme in \trajopt{} is the same as in \ourmethod{}, without the initial seed from our diffusion model.
We use a GPU-accelerated implementation from cuRobo~\cite{sundaralingam2023curobo} and control the computational budget by adjusting optimization iterations.
\item \textbf{\diffuser{}:}
a baseline from Huang et al.~\cite{huang2023diffusion} uses a conditional diffusion model for trajectory planning through sampling. 
Unlike \ourmethod{}, this baseline conditions on point-cloud inputs encoded by Point Transformer~\cite{zhao2021point} instead of a C-space representation. 
We use the author's original network implementation, trained on our dataset.
\item \textbf{Motion Planning Diffusion (\mpd{}):} 
a baseline from Carvalho et al.~\cite{carvalho2023motion} uses a diffusion model for trajectory planning through sampling. 
Unlike \ourmethod{}, \mpd{} employs unconditional diffusion models and relies on sampling guidance for trajectory constraints such as collision avoidance or connecting start and goal configurations. 
We adapt their network to our setup and training dataset. 
\end{itemize}

\paragraph{Comparison to Pure Learning Algorithms}
We consider pure learning algorithms, \diffuser{} and \mpd{},
which neither use a key-configuration environment representation (\claim{1}) nor a motion-planning objective for training diffusion models (\claim{2}).
We evaluate each baseline's diffusion model performance \emph{without} trajectory optimization post-processing, as indicated by the large black, yellow, and purple dots in Figure~\ref{fig:benchmark} corresponding to \ourmethod{}, \diffuser{}, and \mpd{}.
\ourmethod{} consistently outperforms both across all levels. 
For example, in \level{3}, \ourmethod{} achieves a 52.2\% success rate, while \diffuser{} and \mpd{} achieve only 0.6\% and 12.2\%, respectively.

\paragraph{Comparison to Pure Planners}
Compared to \birrt{}, \ourmethod{} uses diffusion-learned trajectory priors to generate collision-free trajectories more efficiently, especially in narrow passages.
In \level{4}, \ourmethod{} achieves a 90\% success rate in 1.0 second, compared to 2.3 seconds for \birrt{}.
Furthermore, as environment complexity increases, 
\birrt{} struggles to find valid segments, widening the success gap from 97.8\% vs. 70.6\% in \level{1} to 90.6\% vs. 46.7\% in \level{4} under a 1-second computational budget.
Next, we consider \trajopt{}, an optimization-based method.
Despite \ourmethod{}'s computational overhead for running the diffusion model, its high-quality initial trajectories lead to faster convergence in complex domains (\claim{3}). 
For example, in \level{2}, \ourmethod{} achieves a 97.2\% success rate in 1.0 second despite an initial overhead of 0.2 seconds, while \trajopt{} remains below 60\% in the same time.
The success rate gap with a 1-second computational budget grows from 26.7\% in \level{1} to 37.3\% in \level{4}, demonstrating \ourmethod{}'s effectiveness in complex environments.

\subsection{Ablation Studies\label{sec:ablation}}
To analyze the impact of our contributions and discuss the claims from Section~\ref{sec:main_results}, we conduct ablation studies using variants of our method, with results shown in Figure~\ref{fig:ablation}. Additional studies are available on our project website.

\begin{itemize}
\item \task{Point-Cloud Conditioning}\textbf{:} To validate \claim{1}, we train a variant of \ourmethod{} conditioned on an equivalent number of workspace point clouds instead of key configurations. We use a patch-based transformer~\cite{yu2021pointbert} to encode the point clouds, keeping the rest of the architecture unchanged.
\item \task{Training Without TrajOpt}\textbf{:} To validate \claim{2}, we train a variant of \ourmethod{} without motion-planning costs (\textit{Collision Loss} and \textit{Distance Loss}). 
The rest of the architecture remains unchanged.
\end{itemize}

\paragraph{Generalization of Key-Configuration Representation (\claim{1})}
Compared to \ourmethod{}, \textit{Point-Cloud Conditioning} exhibits performance degradation across problem levels
and post-processing iterations: collision rates and penetration depth remain higher,
and worsen with increased problem complexity.
For instance, in \level{1-2}, \ourmethod{} outperforms \textit{Point-Cloud Conditioning} by 1.4\% in success rate, 0.3\% in collision rate, and 0.001m in penetration depth. 
In \level{3-4}, the gaps increase to 3.6\%, 1.4\%, and 0.013m.
This highlights the advantage of using C-space representations over point-cloud-based conditioning in complex scenes.

\paragraph{Efficacy of Training with Motion-Planning Costs (\claim{2})}
Compared to \ourmethod{}, \textit{Training Without TrajOpt} exhibits performance degradation across all levels. 
Though less severe than \textit{Point-Cloud Conditioning}, the trend is consistent:
for example, in \level{1-2}, \ourmethod{} outperforms \textit{Training Without TrajOpt} by an average of 1.81\% in success rate and 0.2\% in collision rate.
In \level{3-4}, the performance gap increases to 2.7\% in success rate and 0.7\% in collision rate. 
This shows that incorporating motion-planning costs for training diffusion models improves trajectory quality across various domains.

\paragraph{Efficacy of Post-Processing (\claim{3})}
We observe that applying trajectory optimization during post-processing improves performance across all levels. 
Additionally, we validate that the success of \ourmethod{} is largely due to the high-quality, nearly collision-free initial trajectories obtained from our diffusion model.
As shown in Figure~\ref{fig:ablation} (right), \ourmethod{} in \level{4} outperforms \textit{Point-Cloud Conditioning} by achieving a much smaller penetration depth (0.037 m vs. 0.123 m), despite similar initial success rates (51.1\% vs. 47.2\%), leading to faster convergence during trajectory optimization.
\section{Conclusion}
We present \ourmethod{}, a learning-guided motion planning framework that integrates diffusion-based trajectory sampling with post-processing trajectory optimization.
Incorporating C-space environment representations based on key configurations and a motion-planning training objective, our framework efficiently generates collision-free trajectories in unseen environments.
Simulated experiments demonstrate the efficacy of our framework compared to both diffusion-based planning approaches and conventional motion planning methods.
In this work, we assumed known environment geometry for ground-truth collision states at key configurations.
Future work could extend the framework to handle partial observability of unseen geometries.

\vspace{8pt}
{
\begin{spacing}{1.15}
\footnotesize
\noindent
{\bf Acknowledgements}
This work was partially supported by the Institute of Information \& Communications Technology Planning \& Evaluation (Nos. 2022-0-00311, 2022-0-00612. 2024-00509279, 2019-190075), the National Research Foundation of Korea (No. RS-2024-00359085) funded by the Korean government (MSIT), the National Science Foundation (FRR2145283, EFRI-2318065, FAIN-2019844, NRT-2125858), the Office of Naval Research (N00014-22-1-2204, N00014-24-1-2550, N00014-18-2243), DARPA (TIAMAT program HR0011-24-9-0428, Cooperative Agreement HR00112520004 on Ad Hoc Teamwork), the Army Research Office (W911NF-23-2-0004, W911NF-17-2-0181), Lockheed Martin, and UT Austin's Good Systems grand challenge. Peter Stone serves as the Chief Scientist of Sony AI and receives financial compensation for that role. The terms of this arrangement have been reviewed and approved by the University of Texas at Austin in accordance with its policy on objectivity in research.
\end{spacing}
}

{
% only do this for icra
% \renewcommand{\baselinestretch}{0.985} 
% \renewcommand*{\bibfont}{\small}
% \printbibliography
% }

% % only do this for arxiv
\renewcommand{\baselinestretch}{1.03} 
\renewcommand*{\bibfont}{\footnotesize}
\printbibliography
}
\section* {Appendix}
\label{sec:appendix}

\subsection{Incorporating Guidance During Sampling}

In an unconditional diffusion model, test-time guidance~\cite{zhang2022motiondiffuse, janner2022planning} constrains trajectories to specific environments and start/goal configurations. 
While prior works~\cite{carvalho2023motion,huang2023diffusion} rely on test-time guidance for collision avoidance and endpoint constraints, we use only conditional diffusion models and trajectory optimization for strict constraint satisfaction. 
Here, we also include an ablation study on the complementary use of guidance steps during sampling to enhance motion planning performance.

\paragraph{Guidance Function Implementation}

Algorithm~\ref{algo:guide} describes our procedure for computing cost gradients.
As in \mpd{}~\cite{carvalho2023motion},  the guidance function includes collision and smoothness costs as described in Section~\ref{sec:train}.
Collision costs are computed using cuRobo~\cite{sundaralingam2023curobo}, while smoothness costs employ a Gaussian Process prior~\cite{carvalho2023motion,Barfoot2014BatchCT}.
To ensure stability during guidance, we smooth the sampled trajectory with a Gaussian kernel ($\sigma=4.0$) before computing costs, allowing collision-cost gradients to affect neighboring points (line 1). 
We then compute the clamped collision cost ($d_{\texttt{max}}=0.1$) and the smoothness cost, summing them as $k_{\text{smooth}} c_{\text{smooth}} + k_{\text{coll}}c_{\text{coll}}$, with $k_{\text{smooth}}=1e-9$ and $k_{\text{coll}}=1e-2$ (line 2-4).
Gradients are computed with PyTorch~\cite{paszke2019pytorch}, clamped ($g_{\texttt{max}}=1.0$) to prevent erratic updates, zeroed at endpoints, and directly added to the trajectory (line 5-6).

\begin{algorithm}[b]
\caption{Guidance Step with Cost Gradients.}
\label{algo:guide}
\begin{algorithmic}[1]
\scriptsize
\Require Trajectory $\tau$, Smoothing Kernel $\mathcal{K}$, Environment $\mathcal{G}$
\Ensure Output trajectory $\tau_{\texttt{guide}}$
\State $\tilde{\tau} = \texttt{Convolve}(\texttt{UnNormalize}(\tau), \mathcal{K})$
\Statex // Computing costs
\State $c_{\texttt{coll}} = \texttt{min}(\texttt{CollisionCost}(\tilde{\tau}, \mathcal{G}), d_{\texttt{max}})$
\State $c_{\texttt{smooth}} = \texttt{GPCost}(\tau)$
\State $c_{\texttt{guide}} = k_{\texttt{smooth}}c_{\texttt{smooth}} + k_{\texttt{coll}}c_{\texttt{coll}}$
\Statex // Applying guidance gradients
\State $g_{\texttt{guide}} = \texttt{Clamp}(\nabla_{\tau}(c_{\texttt{guide}}), g_{\texttt{max}})$
\State $\tau_{\texttt{guide}} = \tau - g_{\texttt{guide}}$
\State \Return $\tau_{\texttt{guide}}$
\end{algorithmic}
\end{algorithm}

\begin{figure*}
	\centering
	\includegraphics[width=\linewidth]{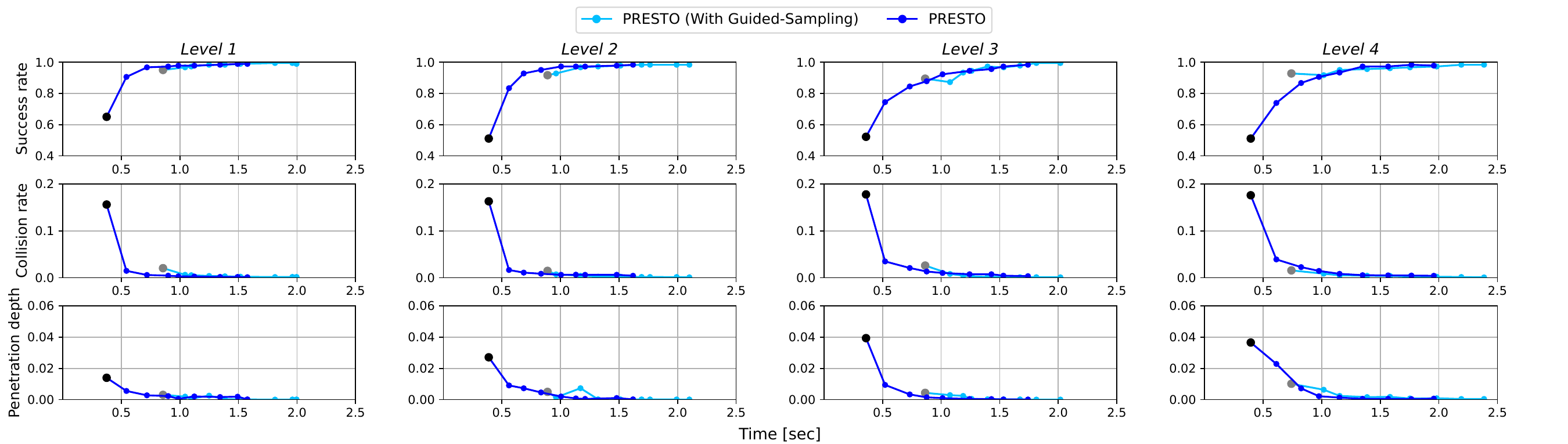}
	\caption
	{
    \textbf{Results with guided-sampling.} We report the success rate (\%), the collision rate (\%), and the penetration depth (m) averaged across 180 problems for \ourmethod{} and the self-variant baselines. Black dots represent PRESTO without post-processing, while gray dots represent PRESTO with Guided Sampling, also without post-processing.
}
    \label{fig:guidance_time}
    \vspace{8pt}
\end{figure*}

\paragraph{Results}
Figure~\ref{fig:guidance_time} compares our model with a variant that includes guidance steps during denoising iterations. 
Overall, guided-sampling enhances the quality of initial trajectories (black dots represent \ourmethod{} without guidance, and gray dots represent \ourmethod{} with guidance).
For example, the success rate of the denoised trajectory before optimization reaches 92.8\% in \level{4}, compared to 51.1\% for \ourmethod{} without guidance. 
However, incorporating guidance requires evaluating cost gradients at each diffusion iteration, resulting in computational overhead. 
In \level{3}, this overhead accumulates to an average of 0.38 seconds, indicating that the additional cost of guidance steps may occasionally degrade performance within a given time frame. 
Despite this, guidance generally improves performance across \level{1-4} in all three metrics: success rate, collision rate, and penetration depth, given the same number of trajectory optimization iterations.

\begin{figure*}
    \vspace{4pt}
	\centering
	\includegraphics[width=\linewidth]{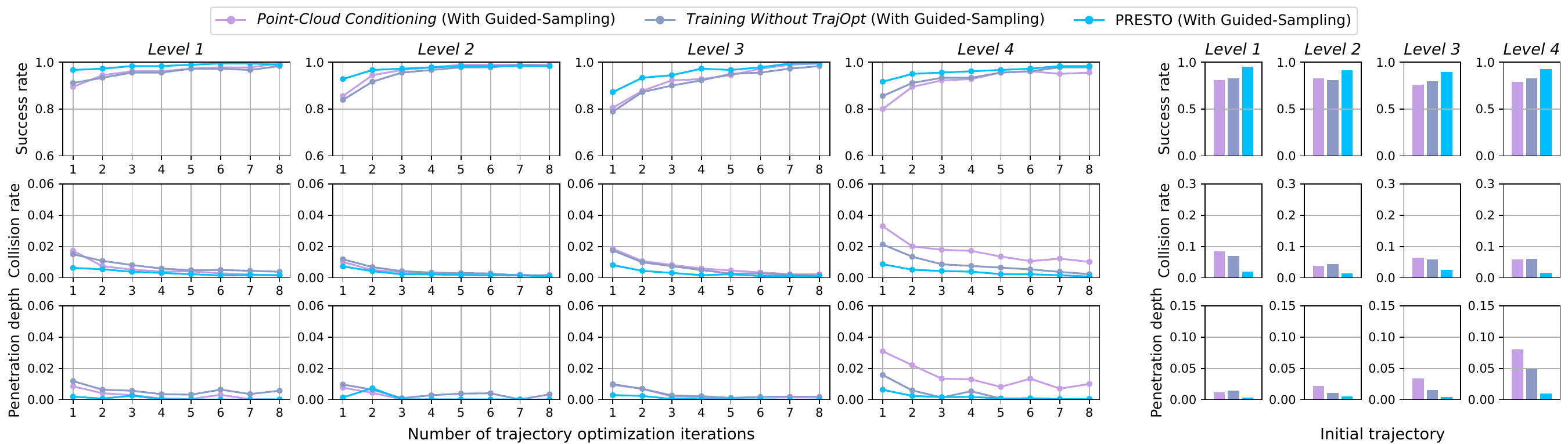}
	\caption
	{
    \textbf{Ablation studies with guided-sampling.} We report the success rate (\%), the collision rate (\%), and the penetration depth (m) averaged across 180 problems for \ourmethod{} and the self-variant baselines with guided-sampling. (\textbf{Left}) We show performance changes with varying post-processing iterations.  (\textbf{Right}) We present the performance of trajectories directly generated by the diffusion model without post-processing.
}
    \label{fig:guidance_ablation}
    \vspace{8pt}
\end{figure*}

We also report the effects of guidance on variants of our model in Figure~\ref{fig:guidance_ablation}. 
Consistent with the results in Figure~\ref{fig:guidance_time}, 
performance across all baselines improves when guidance steps are applied compared to the original results in Figure~\ref{fig:ablation}.
Notably, the success rate gap between \ourmethod{} and its variants widens with the application of guidance. For instance, across \level{1-4}, the gap between \ourmethod{} and the closest baseline (\textit{Point-Cloud Conditioning}) increases from 2.4\% to 4.0\%.
As \ourmethod{} generates higher-quality initial trajectories with smaller penetration depths, spurious collisions are resolved with just a few guidance steps, leading to greater success rate gains compared to the ablations of \ourmethod{}.

\subsection{Computation Time Details}
\begin{figure*}
	\centering
	\includegraphics[width=\linewidth]{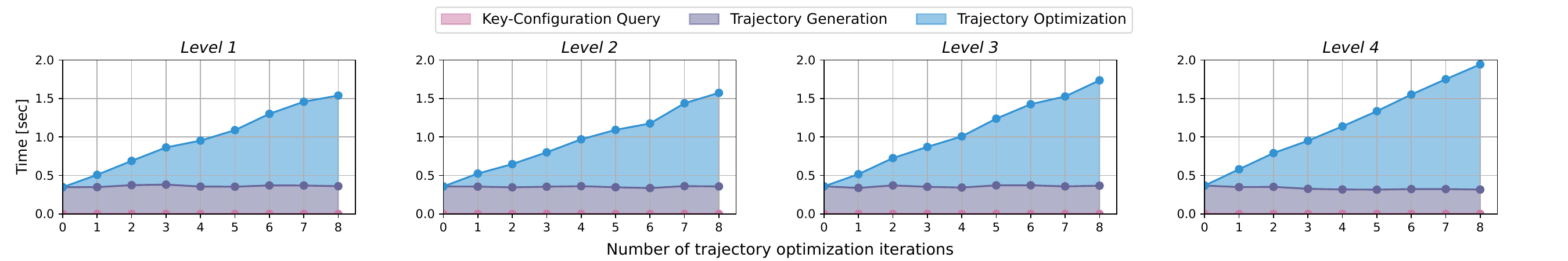}
	\caption
	{
    \textbf{Breakdown of computation time for \ourmethod{}.} We report the average computation time (sec) across 180 problems for \ourmethod{} and provide a breakdown of the time taken for each step: the key-configuration query for environment representation, trajectory generation using the diffusion model, and trajectory optimization during post-processing.}
    \label{fig:breakdown}
\end{figure*}

We provide computation time details for \ourmethod{} in Section~\ref{sec:exp-setup} and present them in Figure~\ref{fig:breakdown}. 
The key-configuration query incurs an average overhead of 0.658 milliseconds, which is negligible compared to trajectory generation in the diffusion model, averaging 0.345 seconds, and trajectory optimization during post-processing, which scales linearly with the number of iterations.
The performance improvements of \ourmethod{}, leveraging an environment representation based on key configurations, over the \textit{Point-Cloud Conditioning} baseline in Section~\ref{sec:ablation} demonstrate its effectiveness in enhancing representation quality without compromising computational speed. 
This makes it particularly well-suited for scenarios with limited computational resources.

\subsection{Point-Cloud Encoder Architecture}

For our ablation with point-cloud inputs in Section~\ref{sec:ablation}, we design the point-cloud encoder based on recent patch-based transformers~\cite{yu2021pointbert,pang2022pointmae}. 
We divide the $\mathbb{R}^{1024\times 3}$ point cloud into 8 patches using farthest-point sampling and k-nearest neighbors ($k=128$). 
Each patch is normalized, flattened, and projected into shape embeddings via a 3-layer MLP with GeLU and Layer Normalization. Positional embeddings, computed from patch centers using a 2-layer MLP, are added before processing with a 4-layer transformer to extract geometric features, which serve as additional input tokens for the DiT in the diffusion model.

\subsection{Background}

\paragraph{Denoising Diffusion Probabilistic Models}
\label{subsec:ddpm}
The diffusion model involves a forward diffusion process, which starts from a perfect solution (in our case a collision-free trajectory) and gradually injects noise until reaching an isotropic Gaussian distribution, and a reverse diffusion process, which learns to generate data through gradual denoising. 
Let $i\in\{0, 1, ..., N\}$ denote a denoising step. The data generation process using a diffusion model involves iteratively denoising noise $\tau_i$ until reaching a denoised sample $\tau_0$, where $\tau$ is a sequence of robot configurations in this work. The reverse diffusion process is given by:
\begin{equation*}
\label{eqn:diffusion}
p_\theta(\tau_{0:T})=p(\tau_T)\prod_{i=1}^N p_\theta(\tau_{i-1}|\tau_i),
\end{equation*}
where $p_\theta(\tau_{i-1}|\tau_i)=\mathcal{N}\big(\tau_{i-1};\mu_\theta(\tau_i, i), \Sigma_\theta(\tau_i, i)\big)$, as proven by Ho~\etal~\cite{ho2020denoising} under some mild conditions. Ho~\etal~\cite{ho2020denoising} show that $\mu_\theta(\tau_i, i)$ can be modeled as a function $\epsilon_\theta(\tau_i, i)$, which predicts a noise value from a sample at the $i$-th denoising step. In addition, they show that the training objective can be simplified $\min_\theta ||\epsilon_\theta(\tau_i, i) - \epsilon_i||^2$, and the diffusion model can be trained by minimizing this term.

\paragraph{Conditional Diffusion Models}
\label{subsec:diffusion}
We convert the unconditional form of \(p_\theta(\tau_{i-1}|\tau_i)\) to be conditioned on our environment representation. We adopt the methods proposed by Dhariwal and Nichol~\cite{dhariwal2021diffusion}, whose objective is to generate an image conditioned on the class. Specifically, they demonstrate that the term of the conditional reverse diffusion process can be factorized as follows: $p_{\theta,\phi}(\tau_{i-1}|\tau_{i}, \phi)\propto p_\theta(\tau_{i-1}|\tau_{i})p_\phi(\phi|\tau_{i-1})$, where $\phi$ is the conditioning variable, which in our case will be an environment representation. Under some mild conditions, they show that after applying the logarithm, we get $\log \big(p_\theta(\tau_{i-1}|\tau_{i})$ $p_\phi(\phi|\tau_{i-1})\big)\approx\log p(u) + \mbox{const}$, where $u\sim\mathcal{N}\big(\tau_{i-1};\mu_\theta(\tau_i, i) + g\Sigma_\theta(\tau_i, i),$ $ \Sigma_\theta(\tau_i, i)\big)$ and $g=$ $\bigtriangledown_{\tau_{i-1}}\log p_\phi(\phi|\tau_{i-1})|_{\tau_{i-1}=\mu_\theta(\tau_i, i)}$. Notice that the mean parameter of the Gaussian term in $u$ includes an additional component, $g\Sigma_\theta(\tau_i, i)$, which is not a part of the mean parameter of $p_\theta(\tau_{i-1}|\tau_i)$. This mathematical derivation implies that the diffusion model can generate a sample conditioned on $\phi$ by the gradients $g$.
Based on this, we particularly use a classifier-free guidance diffusion model to implement our motion generation. It is known that conditional diffusion steps can be run by incorporating the probabilities from both a conditional and an unconditional diffusion model~\cite{ho2022classifier}, as $\bigtriangledown\log p_\phi(\phi|\tau) = \bigtriangledown\log p_\phi(\tau|\phi) - \bigtriangledown\log p_\phi(\tau)$. Therefore, by plugging into the conditional probabilities, conditional guidance can be trained together with the diffusion models, eliminating the need for a separate classifier.

\end{document}